%% file: main.tex
\documentclass[10pt,twocolumn,letterpaper]{article}

\usepackage[pagenumbers]{cvpr} 
\usepackage{algorithm}
\usepackage{algpseudocode}
\usepackage{amsmath}
\usepackage{amssymb}
\usepackage{multirow}
\usepackage{makecell}
\usepackage{xcolor}

\input{preamble}

\definecolor{cvprblue}{rgb}{0.21,0.49,0.74}
\usepackage[pagebackref,breaklinks,colorlinks,allcolors=cvprblue]{hyperref}

\def\paperID{*****} 
\def\confName{CVPR}
\def\confYear{2025}

\title{\method:  Learning Negative Embedding with Reward Guidance}

\author{
Xiaomin Li$^{1,2,*}$,
Yixuan Liu$^{1,*,\ddagger}$,
Takashi Isobe$^{1}$,
Xu Jia$^{2,\dagger}$,
Qinpeng Cui$^{3}$,
Dong Zhou$^{1}$,\\
Dong Li$^{1}$,
You He$^{3}$,
Huchuan Lu$^{2}$,
Zhongdao Wang$^{3,\dagger}$,
Emad Barsoum$^{1}$\\
$^1$Advanced Micro Devices Inc., $^2$Dalian University of Technology, $^3$Tsinghua University\\
\tt\small xmli22@mail.dlut.edu.cn, \{yixuan.liu,takashi.isobe,dong.zhou,d.li,ebarsoum\}@amd.com, \\
\tt\small \{xjia,lhchuan\}@dlut.edu.cn, heyou\_f@126.com, \{cqp22,wcd17\}@tsinghua.edu.cn
}

\begin{document}
\twocolumn[{%
\renewcommand\twocolumn[1][]{#1}%
\maketitle
\begin{center}
    \centering
    \captionsetup{type=figure}
    \includegraphics[width=1\textwidth]{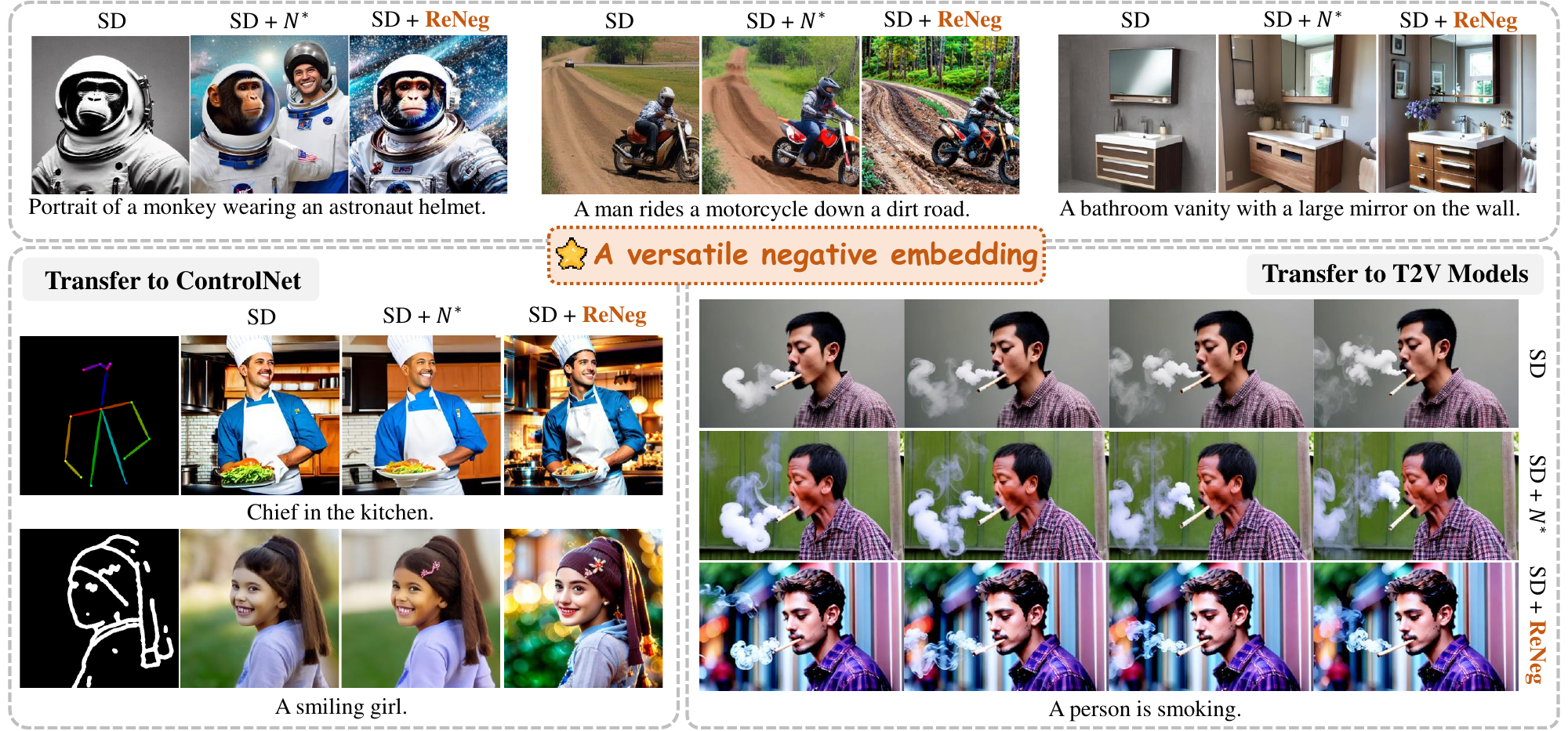}
    \vspace{-0.6cm}
    \captionof{figure}{We develop ReNeg, a versatile negative embedding seamlessly adaptable to text-to-image and even text-to-video models. Strikingly simple yet highly effective, ReNeg amplifies the visual appeal of outputs from base Stable Diffusion~(SD) models. ‘$+N^*$’ and ‘+ReNeg’ indicate improved results with handcrafted negative prompts and our negative embedding, respectively.}
    \label{fig:teaser}
\end{center}%
}]

\let\thefootnote\relax\footnotetext{$^*$ Equal contribution. Work done during an internship at AMD.} 
\let\thefootnote\relax\footnotetext{$^\dagger$ Corresponding authors.
}
\let\thefootnote\relax\footnotetext{$^\ddagger$ Project lead.
}

\input{sec/0_abstract}    
\input{sec/1_intro}
\input{sec/2_related_work}
\input{sec/3_method}
\input{sec/4_experiments}

\input{sec/5_conclusion}
{
    \small
    \bibliographystyle{ieeenat_fullname}
    \bibliography{main}
}

\end{document}

%% file: preamble.tex
\newcommand{\method}{ReNeg\xspace}

%% file: sec/0_abstract.tex
\begin{abstract}
In text-to-image (T2I) generation applications, negative embeddings have proven to be a simple yet effective approach for enhancing generation quality. 
Typically, these negative embeddings are derived from user-defined negative prompts, which, while being functional, are not necessarily optimal.
In this paper, we introduce \textbf{\method}, an end-to-end method designed to \emph{learn} improved \textbf{Neg}ative embeddings guided by a \textbf{Re}ward model. 
We employ a reward feedback learning framework and integrate classifier-free guidance (CFG) into the training process, which was previously utilized only during inference, thus enabling the effective learning of negative embeddings.
We also propose two strategies for learning both global and per-sample negative embeddings. 
Extensive experiments show that the learned negative embedding significantly outperforms null-text and handcrafted counterparts, achieving substantial improvements in human preference alignment. 
Additionally, the negative embedding learned within the same text embedding space exhibits strong generalization capabilities.
For example, using the same CLIP text encoder, the negative embedding learned on SD1.5 can be seamlessly transferred to text-to-image or even text-to-video models such as ControlNet, ZeroScope, and VideoCrafter2, resulting in consistent performance improvements across the board.
Code and learned negative embeddings are released in \hyperlink{https://github.com/AMD-AIG-AIMA/ReNeg}{this URL}.

\end{abstract}

%% file: sec/1_intro.tex
\section{Introduction}
\label{sec:intro} 
Recent advancements in diffusion models~\cite{ddpm,score-based,ddim} have led to significant breakthroughs in image generation~\cite{guided-diffusion,ldm,fastcomposer} and video generation~\cite{tune-a-video,svd,show-1,vc2,motrans}. A pivotal technique in these models is Classifier-free Guidance (CFG)~\cite{cfg}, which enhances text control capabilities while also improving the realism and aesthetic quality of the generated outputs. CFG operates by concurrently training a conditional probability model alongside an unconditional probability model, merging their score predictions during inference. The unconditional model takes a null-text prompt as input, serving as guidance for the score predictions. In practice, this null-text prompt can be substituted with a negative prompt~\cite{dpo-diff,dnp,Impact_of_Negative_Prompts,woolf2022}, prompting the model to generate outputs that deviate from the characteristics specified by the negative prompt. Empirical evidence suggests that negative prompts generally yield superior results compared to null-text prompts, leading to their prevalent use in text-to-image~\cite{dpo-diff,dnp} and text-to-video generation~\cite{animatediff}.

To create effective negative prompts, the prevailing method involves manually selecting negative terms—such as ``low resolution'' and ``distorted''—and employing trial and error to identify optimal combinations. However, this approach suffers from significant limitations. The manually defined search space is inherently incomplete, as it cannot encompass all permutations of negative vocabulary. Additionally, the quality assessment of generated outputs often relies on subjective human judgment, resulting in inefficiencies during the search process. Consequently, manually crafted negative prompts, while yielding decent performance, are suboptimal.

In this work, we introduce \method, a \textbf{Re}ward-guided approach that directly \emph{learns} \textbf{Neg}ative embeddings through gradient descent. Our method enhances the process along two dimensions compared to manual searches. First, learning occurs within a comprehensive search space. We utilize the continuous text embedding space—specifically, the output embedding space of text encoders—since the original language space is discrete~\cite{dpo-diff}, which makes gradient-based optimization more challenging. Second, we fully automate the evaluation criterion using an image reward model (RM)~\cite{imagereward}, enabling continuous gradient descent rather than relying on trial and error.
We formulate the learning objective within a reward feedback framework, leveraging a pretrained image reward model to guide the gradient descent. Negative embeddings are treated as a set of model parameters, and to ensure they receive gradients from the reward guidance, we modify the training process to incorporate CFG, which was previously utilized only during inference. Furthermore, we propose a tuning approach that learns per-sample negative embeddings, adapting to various prompts and yielding additional improvements.

Extensive experiments demonstrate that both the learned global and per-sample negative embeddings consistently outperform null-text and carefully crafted negative embeddings in terms of generation quality and human preference. 
In human preference benchmarks such as HPSv2~\cite{hpsv2} and Parti-Prompts~\cite{partiprompts}, \method, while significantly simpler, achieves remarkable results, even rivaling methods that require full model fine-tuning.
Moreover, we highlight that the global negative embeddings learned using \method exhibit\emph{ strong generalization capabilities} and are \emph{easily transferable}.
Any text-conditioned generative model utilizing the same text encoder can share these negative embeddings.
We demonstrate that negative embeddings learned with Stable Diffusion 1.5 transfer seamlessly to text-to-image and text-to-video models, including ControlNet~\cite{controlnet}, ZeroScope~\cite{zeroscope}, and VideoCrafter2~\cite{vc2}.

Our primary contributions are summarized as follows:
\begin{itemize}
    \item We propose \method, an innovative approach for learning negative embeddings guided by reward models.
    \item We introduce two strategies for learning both global and per-sample negative embeddings, each yielding significant improvements, with per-sample embeddings demonstrating superior performance.
    \item We establish that the learned global negative embeddings can be readily transferred to other models and tasks, resulting in consistent performance improvements.
\end{itemize}







%% file: sec/2_related_work.tex
\section{Related Works}
\label{sec:related_works}
\noindent
\textbf{Text-to-Image Diffusion Models.}
Existing T2I models~\cite{ldm,imagen} have demonstrated impressive generative capabilities, but generating satisfactory content from user-provided positive text prompts remains challenging, particularly for common issues such as hand and face generation defects. A straightforward solution is to improve the user prompt through prompt engineering~\cite{systematic,prompt_engineering}. Recent works~\cite{promptist,beautifulprompt} have optimized pretrained large language models~(LLMs)~\cite{bloom,gpt2}, enabling the modified models to generate refined prompts based on the original user input. While the image quality generated using these refined prompts shows some improvement over the original, exploration of negative prompts remains limited. DNP~\cite{dnp}, for instance, samples a negative image based on the positive prompt and uses a captioning model to generate a corresponding negative prompt. In practice, this process often requires multiple attempts to find an appropriate negative prompt. On the other hand, DPO-Diff~\cite{dpo-diff} searches for negative prompts within the discrete language space. In contrast, we attempt to directly learn negative embeddings through gradient descent in the continuous text embedding space, which allows for a more efficient search of optimal negative embedding.

\noindent
\textbf{Reward Optimization for Text-to-Image Models.}
Numerous efforts~\cite{imagereward,textcraftor,draft} have been made to optimize existing T2I models by leveraging reward models~\cite{hpsv2,Laion-5b,pick-a-pic,imagereward}. Inspired by Reinforcement Learning from Human Feedback (RLHF)~\cite{deep_reinforcement,human_feedback} in LLMs, DDPO~\cite{ddpo} employs reinforcement learning~(RL) to finetune diffusion models, aiming to maximize the reward score within a relatively constrained vocabulary. Alternatively, Diffusion-DPO~\cite{diffusion-dpo} adopts the Direct Preference Optimization~(DPO)~\cite{dpo} strategy that aligns diffusion models with human preferences without the need for RL, significantly improving the visual appeal of generated content. 
Moreover, methods like ReFL~\cite{imagereward} and TextCraftor~\cite{textcraftor} calculate the reward score directly from the predicted initial image to guide T2I model finetuning, typically targeting components such as the text encoder~\cite{textcraftor} or UNet~\cite{diffusion-dpo,ddpo,imagereward} in the diffusion model. Distinct from these approaches, our method seeks to optimize the negative embedding under reward guidance, enabling comparable or even superior generation quality at minimal storage cost.

%% file: sec/3_method.tex
\section{Preliminary}
\noindent
\textbf{Diffusion Models.}
Diffusion models consist of two processes: a forward noising process and a backward denoising process. In the forward process, Gaussian noise is gradually added to the data $x_0\sim p(x_0)$ through a fixed-length Markov chain. As the time steps increase,  a series of noisy latent variables $\{x_1, x_2, ..., x_T\}$ with increasing noise levels are progressively generated:
\begin{equation}
    q(x_t|x0) = \mathcal{N}(x_t; \sqrt{1-\bar{\beta}_{t}}x_0, \bar{\beta}_t\textbf{\textit{I}}),
\end{equation}
\begin{equation}
    x_t = \sqrt{\bar{\alpha}_t}x_0 + \sqrt{1-\bar{\alpha}_t}\epsilon,
\end{equation}
where $\alpha_t = 1 - \beta_t$ and $\bar{\alpha}_t = \prod_{i=1}^t\alpha_i$. $\epsilon$ is a standard Gaussian noise, and $\alpha_t$ decreases with the timestep $t$. In the backward process, diffusion models restore the original data distribution by progressively denoising variables sampled from a Gaussian distribution $x_T \sim \mathcal{N}(0, \textbf{\textit{I}})$. 
\begin{equation}
    p_\theta(x_{t-1}|x_t)=\mathcal{N}(x_{t-1}; \mu_\theta(x_t, t), \Sigma_\theta(x_t, t)),
\end{equation}
$\mu_\theta$ and $\sigma_\theta$ are predicted statistics. Observed by~\cite{ddpm}, only predicting the noise through a neural network $\epsilon_\theta(x_t, t)$ works well. For diffusion-based text-to-image tasks, the textual prompt $c$ is introduced as the condition. The training objective can be represented by a reconstruction loss:
\begin{equation}
    \mathcal{L}_{\text{rec}} = \mathbb{E}_{\mathbf{x_0}, \mathbf{c}, \boldsymbol{\epsilon} \sim \mathcal{N}(0, \textbf{\textit{I}}), t} \left[\lVert \boldsymbol{\epsilon}_\theta(x_t, E(c), t) - \boldsymbol{\epsilon} \rVert_2^2 \right].
\end{equation}
Here, $E$ is a pre-trained text encoder.

\noindent
\textbf{Classifier-free Guidance.}
During training diffusion models, \cite{cfg} proposes to jointly train a conditional diffusion model $\boldsymbol{\epsilon}_\theta(x_t, c, t)$ and an unconditional one $\boldsymbol{\epsilon}_\theta(x_t, \phi, t)$.
During inference, it allows control over the balance between realism and diversity of generated samples by adjusting the guidance scale $\gamma$ for the conditional generation task:
\begin{equation}
\label{eq:cfg}
    \tilde{\epsilon}_\theta(x_t, c, t)=\boldsymbol{\epsilon}_\theta(x_t, \phi, t)+\gamma(\boldsymbol{\epsilon}_\theta(x_t, c, t)- \boldsymbol{\epsilon}_\theta(x_t, \phi, t))
\end{equation}
The unconditional model is realized by inputting a null-text embedding $\phi$ as condition. Numerous works~\cite{dpo-diff,dnp} indicate that replacing $\phi$ with a handcrafted negative embedding $n$ further improves generation quality. 
In this work, we aim to learn the negative embedding. The main challenge here is that CFG is usually performed during inference only. To learn the negative embedding, we have to incorporate CFG into training and make the gradient \wrt $n$ tractable.

\section{Method}
\label{sec:method}
In this section, we begin by discussing our motivation and the feasibility of learning a negative embedding in Section~\ref{sec:feasibility}, illustrating why it could possibly work. 
Next,  we introduce how to learn a universal negative embedding using \method in Section~\ref{sec:method_global},
and present a per-sample variant in Section~\ref{sec:method_adaptive} to further enhance image generation quality.

\begin{figure*}[!t]
\centering
\includegraphics[width=1\textwidth]{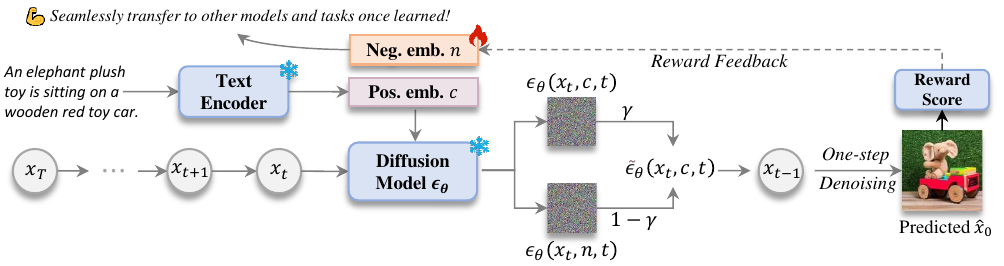}
\caption{Overview of the training pipeline of our~\method. We learn the negative embedding by integrating Classifier-Free Guidance into the training process. The negative embedding is directly optimized using reward feedback, with gradients backpropagated during the final one-step generation process. Once optimized, it can be seamlessly transferred to any T2I or T2V models.}
\label{fig:overall_framework}
\end{figure*}

\begin{table}[t]
\renewcommand\arraystretch{1.1}
\centering
\caption{
Parameter efficiency comparison between the negative embedding $n$, the full model parameters $\theta_0$, and the added LoRA parameters with rank $r=8$ ($\theta_{l}^{8}$) and $r=16$ ($\theta_{l}^{16}$).
}
\label{tab:pilot}
\resizebox{0.38\textwidth}{!}{
    \begin{tabular}{cccccc}
    \toprule   {$E(n)$} & { $E(\theta_0)$}    & {$E(\theta_{l}^{8})$} & {$E(\theta_{l}^{16})$}  \\ 
    \midrule
    $5.1\times 10^{-4}$ & $1.5\times10^{-6}$  & $8.1\times 10^{-8}$ &$1.9\times 10^{-9}$ \\
    
    \bottomrule
    \end{tabular}
}
\end{table}

\subsection{On feasibility of learning negative embeddings}
\label{sec:feasibility}
Our key insight is that the negative embedding can be seen as part of the model parameters.
Based on a pretrained diffusion model, we make an important observation that \emph{tuning the negative embedding is much more efficient than tuning the model parameters}. 
To illustrate this, we conduct a pilot study as follows: we randomly select a set of $N$ prompts, feed them into a pretrained SD1.5 model, and obtain a set of denoised latents ${X}\in\mathbb{R}^{N\times D}$, where $D = h\times w \times c$ represents the dimension of the latents. 
When $N$ is large enough, ${X}$ can be seen as a nice sampling of the learned distribution.
We care about how the distribution changes when small perturbations are applied to the model parameters $\theta$ (or similarly to the negative embedding $n$).
For a desired ``efficient'' set of parameters, we expect the distribution to vary fast with small perturbations, therefore it would bring a greater chance to reach better solutions when tuning this set of parameters.
To quantify parameter efficiency, we define a criterion based on the {Jacobian matrix}.
Let $\mathcal{J} = \frac{\partial X}{\partial \theta} \in \mathbb{R}^{d_\theta \times ND}$ be the Jacobian matrix, computed via iterative backpropagation (See Appendix for details), where $d_\theta$ is the dimension of parameter $\theta$. 
We define the parameter efficiency $E(\theta) = \frac{1}{NDd_\theta} || \mathcal{J}||_F$, where $||\cdot||_F$ denotes the Frobenius Norm of a given matrix. A larger $E(\theta)$ implies a higher rate of distributional change, indicating that tuning the parameters $\theta$ is more efficient. 

We compare the parameter efficiency between the full model parameters $\theta_0$, the added LoRA~\cite{lora} parameters with rank $r=8$ ($\theta_{l}^{8}$) and $r=16$ ($\theta_{l}^{16}$), and the negative embedding $n$.
We observe in Table~\ref{tab:pilot} that $E(n)  \gg E(\theta_0)  > E(\theta_{l}^{8}) > E(\theta_{l}^{16})$, indicating that tuning the negative embedding is the most efficient. The lower efficiency of $\theta_0$, $\theta_{l}^{8}$ and $\theta_{l}^{16}$ can be attributed to the fact that they start from a well-pretrained checkpoint where local minimum has already been reached, leading to a rather low rate of distributional change. In contrast, the negative embedding is usually set empirically and is thus far from optimal. The high parameter efficiency shows tuning the negative embedding is a promising alternative to tuning the entire model, highlighting the feasibility of \emph{learning} negative embeddings. 

\subsection{Learning Negative Embeddings with \method}
\label{sec:method_global} 
\noindent
\textbf{Overview.} The key idea behind \method is to learn the negative embedding using a reward feedback learning (ReFL) framework~\cite{imagereward}. Starting with a pretrained diffusion model, \eg, Stable Diffusion 1.5~\cite{ldm}, we sample random prompts and feed them into the model to generate predicted images. An image reward model evaluates the reward score according to the generation quality, and then backpropagate gradients to the diffusion model. A graphical overview of this process is provided in Fig.~\ref{fig:overall_framework}.

\noindent
\textbf{Reward model.}
In language models, reinforcement learning from human feedback (RLHF)~\cite{deep_reinforcement,human_feedback} is a widespread technique that is proven to be effective in human preference alignment. Typically, the reward for a given prediction is evaluated by a pretrained reward model (RM) $\mathcal{R}$. In the domain of visual data generation, we adopt a similar approach, leveraging an RM to guide diffusion models. 
Specifically, using the human preference-based reward model HPSv2.1~\cite{hpsv2}, we optimize the negative embedding to maximize the reward score. 
Formally, an image RM outputs a scalar reward score $r = \mathcal{R}(c,x)$ for a given pair of prompt $c$ and image $x$. 

\noindent
\textbf{Learning objective.}
Unlike in language models, the iterative denoising nature of diffusion models makes it challenging to estimate the likelihood of the generated samples using reward models. Specifically, the reward model is typically trained on natural images and struggles to estimate the rewards of intermediate images that are not fully denoised.
A feasible solution is to perform a one-step prediction from intermediate denoised latents and then supervise the resulting prediction. Concretely, suppose a canonical denoising iteration is $x_T \rightarrow \ldots \rightarrow x_{t} \rightarrow \ldots \rightarrow x_0 $. At an intermediate timestep $t$, we directly predict $\hat{x}_0$ from $x_t$, \ie, the iteration can be depicted as $x_T \rightarrow \ldots \rightarrow x_{t} \rightarrow \hat{x}_0 $. The one-step prediction is given by
\begin{equation}
\label{eq:x_0_pred}
    \hat{x}_0 = \frac{x_t - \left( \sqrt{1 - \bar{\alpha}_t} \, \boldsymbol{\epsilon}_{\theta}(x_t, c, t) \right)}{\sqrt{\bar{\alpha}_t}}.
\end{equation}
The final learning objective is to maximize the expectation of reward scores over the prompt distribution $\mathcal{D}$: 
\begin{equation}
\label{eq:obj}
    \mathcal{J}_\theta (\mathcal{D}) = \mathbb{E}_{c\sim \mathcal{D}}{(\mathcal{R}(c, \hat{x}_0))}.
\end{equation}
For implementation simplicity, the gradient is backpropagated through $\hat{x}_0$ to $x_t$, but stops to go through further to $x_{t+1}, \ldots, x_T$. \cite{imagereward} reveal that the selection of the intermediate timestep $t$ is non-trivial. When $T-t$ is small, rewards for all generations remain indistinguishably low. When $T-t$ is sufficiently large, rewards for generations of different quality become distinguishable. To ensure efficient training, we set $T=30$ and randomly sample $t\in [0,10]$.

\noindent
\textbf{Training with CFG.} In order to learn negative embedding, we incorporate CFG into the aforementioned training framework.
We register the negative embedding as a set of the model parameters and initialize it with the pre-extracted null-text embedding. During training, all other parameters are frozen, and only the negative embedding is updated using gradient $\frac{\partial \mathcal{J}_\theta(\mathcal{D})}{\partial n}$. 
We also experimented with making the guidance scale factor $\gamma$ learnable, but found similar results to setting it as a constant. Therefore, we opt to keep it as a constant.
Furthermore, effective alignment between training and inference processes can be achieved through this CFG training strategy, where the predicted noise at each step is reparameterized using Eq.~\ref{eq:cfg} during both processes. The sample $x_{t-1}$ is then predicted from the latents of the previous timestep and the reparameterized noise.

\begin{figure}[t]
\begin{center}
\includegraphics[width=1\linewidth]{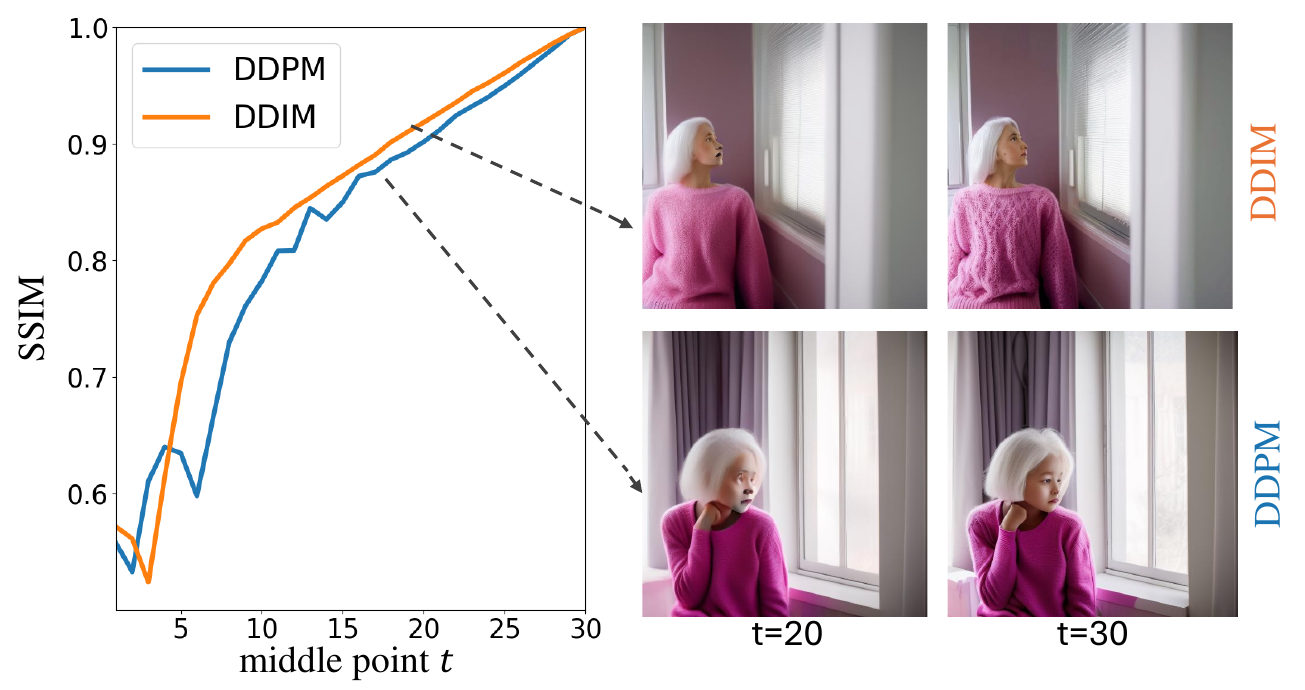}
\end{center}
   \vspace{-0.5cm}
   \caption{Deterministic ODE sampler (DDIM) improves $\hat{x}_0$ prediction. 
   We plot the similarity score between $\hat{x}_0$ and $x_0$  against varying selection of middle point $t$. DDIM consistently outperforms DDPM. \textbf{Prompts:} \emph{A white-haired girl in a pink sweater looks out a window in her bedroom. }} 
   
\label{fig:ddim}
\end{figure}

\noindent
\textbf{Deterministic ODE solver improves $\hat{x}_0$ prediction.}
Due to the nature of reward guidance being applied to the one-step prediction $\hat{x}_0$, we aim for the prediction of $\hat{x}_0$ to be as accurate as possible. This accuracy allows for a broader sampling range at time $t$, leading to a more precise learned marginal distribution. We found that, in this context, selecting the deterministic ODE solver of DDIM for sampling from $T$ to $t$ yields better results than using the original SDE solver of DDPM, resulting in more stable and accurate predictions of $\hat{x}_0$. In Figure~\ref{fig:ddim}, we visualize the differences between $\hat{x}_0$ predicted from $x_t$ and $x_0$ obtained through complete sampling. As $t$ varies within the interval $[0,T]$, it is evident that the discrepancies between $\hat{x}_0$ and $x_0$ with DDIM are generally smaller than those with DDPM. 

\noindent
\textbf{Transferability.}  Although our training relies on the generative model and the reward model, the learned negative embedding is independent of these models, as they are merely vectors in the output space of the text encoder. This means that as long as the same text encoder is used, the learned negative embedding can be seamlessly transferred across different generative models. We validate the effectiveness of this transfer in the experimental section. 

\subsection{Per-sample Negative Embedding}
\label{sec:method_adaptive}
While it is feasible to learn a globally applicable and effective negative embedding, the optimal negative embedding may vary depending on the prompt. For instance, when the prompt requests realistic images resembling photography, "cartoonish" may serve as a negative prompt. Conversely, when the prompt asks for a cartoon image, "realism" may instead function as the negative prompt. Therefore, we propose an \emph{optional} procedure that allows further adaptation of the learned global negative embedding to generate adaptive negative embedding tailored to a specific prompt.

\begin{figure}[t]
\begin{center}
\includegraphics[width=1\linewidth]{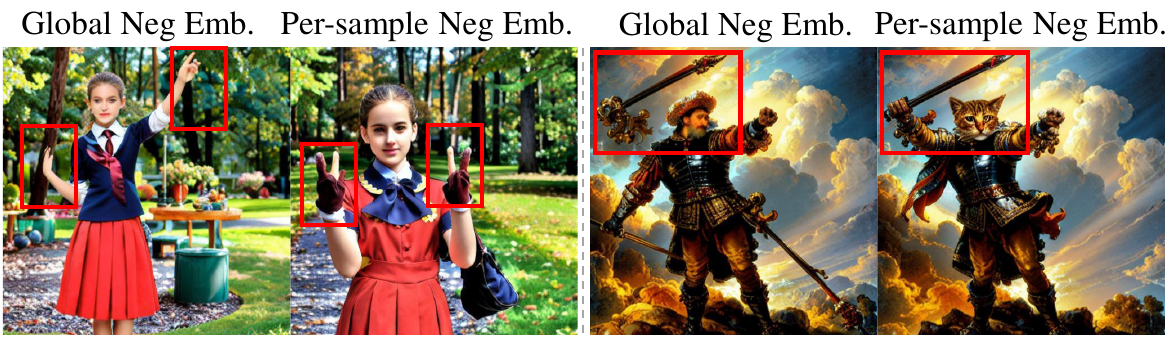}
\end{center}
   \vspace{-0.5cm}
   \caption{Comparison of results using global negative embedding and per-sample negative embedding. Red boxes highlight the improvement in image details achieved by the per-sample negative embedding. \textbf{Prompts: (Left)} \emph{A girl in a school uniform \textcolor{orange}{making a scissor hand gesture}.} \textbf{(Right)} \emph{An oil painting of \textcolor{orange}{a muscular cat} wielding a weapon with dramatic clouds in the background.}}
\label{fig:adaptive_neg_emb}
\end{figure}

Given a specific prompt, our method for learning the per-sample negative embedding is similar to that described in Eq.~\eqref{eq:obj}, with several key differences. First, we no longer need to take the expectation over the prompt distribution, instead training on individual samples. Second, the optimization of per-sample negative embeddings is initialized with the learned global negative embedding. Finally, we propose a search strategy that guarantees convergence to a solution that outperforms the global negative embedding.
Specifically, we define a maximum training step $N$ and a patience value $P$, training stops when all $N$ steps finish, or if early-stop is triggered when the reward does not increase for at least $P$ consecutive steps.
Implementation details are provided in Algorithm~\ref{algo}. We observe consistent improvements, particularly in details refinement and text-image alignment, when training per-sample negative embedding compared to the learned global negative embedding, as shown in Fig.~\ref{fig:adaptive_neg_emb}.

\begin{algorithm}[t]
\caption{Learning per-sample negative embedding}
\label{algo}
\begin{algorithmic}
\Require Prompt $c$, learned global negative embedding~$n$, maximum steps $N$, patience value $P$
\State Variable $\mathcal{J}_{best} \gets 0$,  $p_{ctr} \gets 0$
\For{$i = 1$ to $N$} 
    \State $x_T \sim \mathcal{N}(0, I)$ \quad \quad \quad \quad  \; // \text{Sample noise as latent}
    \State $\hat{x}_0 = \text{Sample} (x_T, c, n)$  \quad // \text{Sampling using Eq.~\eqref{eq:x_0_pred}} 
    \State $\mathcal{J}_n(c) = \mathcal{R}(c, \hat{x}_0)$
    \If{$\mathcal{J}_n(c) > \mathcal{J}_{best}$} \quad 
        \State $\mathcal{J}_{best} \leftarrow \mathcal{J}_n(c)$
        \State Reset $p_{ctr} \gets 0$
    \Else \quad \quad\quad\quad\quad \quad\quad \qquad // \text{Early stopping} 
        \State $p_{ctr} \leftarrow p_{ctr} + 1$
        \If{$p_{ctr} \geq P$} \textbf{break}
        \EndIf
    \EndIf
    \State Update $n$ using gradient descend.
\EndFor
\State Return per-sample negative embedding $n$. 
\end{algorithmic}
\end{algorithm}

%% file: sec/4_experiments.tex
\begin{table*}[t]
\renewcommand\arraystretch{1.1} 
\centering
\caption{Quantitative results on HPSv2 and Parti-Prompts benchmarks. ‘$+N^*$’ represents the handcrafted negative prompt. Methods marked with~$\dagger$ indicate our reproduced version. The best result is highlighted in \textbf{bold}, and the second best is \underline{underlined}.
}
\label{tab:quantitative_results}
\resizebox{0.92\textwidth}{!}{%
\fontsize{26pt}{30pt}\selectfont
\setlength{\tabcolsep}{16pt}
\begin{tabular}{llccccccc}
\toprule
& \multirow{2}{*}{\textbf{Model}} & \multicolumn{4}{c}{\textbf{HPSv2}} & \multicolumn{3}{c}{\textbf{Parti-Prompts}} \\
\cmidrule(lr){3-6} \cmidrule(lr){7-9}
& & Animation  &  Concept Art  & Painting & Photo & Pickscore & Aesthetic & HPSv2.1  \\ 
\midrule
\multirow{2}{*}{Direct Inference} & SD1.5 & 25.92 & 24.66 & 24.65 & 25.62 & 18.40 & 5.23 & 25.67 \\
& SD1.5 + $N^*$ & 27.29 & 26.33 & 26.39 & 27.01 & 19.14 & 5.26 & 26.79 \\
\midrule
\multirow{3}{*}{Prompt Refinement}  & BeautifulPrompt~\cite{beautifulprompt} & 24.07 & 23.95 & 23.99 & 21.40 & 19.38 & \underline{5.78} & 22.72 \\
& Promptist~\cite{promptist} & 26.22 & 25.11 & 25.14 & 24.25 & 19.44 & 5.42 & 25.24 \\
& $\text{DNP}^\dagger$~\cite{dnp} & 26.02 & 25.08 & 24.89 & 25.49 & 19.81 & 5.21 &  25.83 \\
\midrule
\multirow{5}{*}{Finetuning SD} & DDPO-Aesthetic~\cite{ddpo} & 18.20 & 19.03 & 19.15 & 18.93 & 19.23 & 4.99 & 20.69 \\
& DDPO-Alignment~\cite{ddpo} & 20.45 & 20.53 & 20.12 & 20.33 & 19.29 & 4.93 & 19.00 \\
& $\text{Diffusion-DPO}^\dagger$~\cite{diffusion-dpo} & 27.60 & 26.42 & 26.36 & 26.32 & 19.48 & 5.26 & 26.62 \\
& $\text{ReFL}^\dagger$~\cite{imagereward} & 29.04 & 28.34 & 28.21 & 27.48 & 18.17 & 5.48 & 27.97 \\
& TextCraftor~(Text)~\cite{textcraftor} & 30.16 & 30.48 & 30.46 & 28.36 & 19.17 & \textbf{5.90} & 28.36 \\
\midrule
\multirow{2}{*}{ReNeg} & Global Neg. Emb. & \underline{31.37} & \underline{31.67} & \underline{32.00} & \underline{29.27} & \underline{19.90} & 5.45 & \underline{29.16} \\ 
& Per-sample Neg. Emb. & \textbf{32.21} & \textbf{32.52} & \textbf{32.83} & \textbf{30.00} & \textbf{19.97} & 5.50 & \textbf{29.84} \\
\bottomrule
\end{tabular}
}
\end{table*}

\section{Experiments}
\label{sec:experiments}
\subsection{Implementation Details}
\textbf{Dataset.} 
In this work, we use the prompts provided by ImageReward~\cite{imagereward} for training. The training set consists of 10,000 prompts spanning various categories, including people, art, and outdoor scenes. We evaluate the proposed ReNeg on two popular benchmarks: Parti-Prompts~\cite{partiprompts} and HPSv2~\cite{hpsv2}. Parti-Prompts contains 1,632 prompts encompassing various categories. Meanwhile, HPSv2 comprises 3,200 prompts, covering four styles of image descriptions: animation, concept art, paintings, and photo.

\noindent
\textbf{Training Setting.}
The proposed method is built upon the open-sourced~\textit{Stable Diffusion 1.5}.
To optimize the negative embedding, we employ the AdamW~\cite{adamw} optimizer and a Cosine Scheduler for 4,000 steps, with a learning rate of $5e-3$ and batch size of 64. The weights of the pretrained T2I diffusion model are frozen throughout the optimization. We further refine the negative embedding for an additional 10 steps with a patience value of 3 to obtain the per-sample negative embedding.
At inference, DDIM scheduler with 30 steps is used for sampling and the classifier-free guidance weight is set to 7.5 with the resolution 512 × 512. 

\noindent
\textbf{Evaluation metrics.}
We adopt the Human Preference Score v2.1 (HPSv2.1)~\cite{hpsv2}, PickScore~\cite{pick-a-pic}, and an aesthetic predictor~\cite{Laion-5b} to 
comprehensively evaluate our method. Both HPSv2.1 and PickScore are derived from CLIP-based models trained on large-scale human preference datasets, allowing them to approximate human perception in assessing image quality. 
These metrics have demonstrated strong alignment with actual human preferences. 
The aesthetic predictor assesses visual appeal by analyzing high-dimensional image features, focusing on aspects of style and semantics. To further evaluate the performance of existing T2V models enhanced by ReNeg, we employ four metrics from VBench~\cite{vbench}: Aesthetic Quality, Motion Smoothness, Temporal Flickering, and Background Consistency.

\subsection{Comparison with State-of-the-Arts
}
We compare \method with three categories of methods:
(1) Null-text embedding and handcrafted negative embedding; (2) Prompt refinement methods that utilize automatic or manual recaptioning for either positive or negative prompts; and (3) Human preference alignment methods that typically involve tuning all model parameters.  The quantitative results are presented in Tab.~\ref{tab:quantitative_results}.

\begin{figure*}[t]
\begin{center}
\includegraphics[width=1\linewidth]{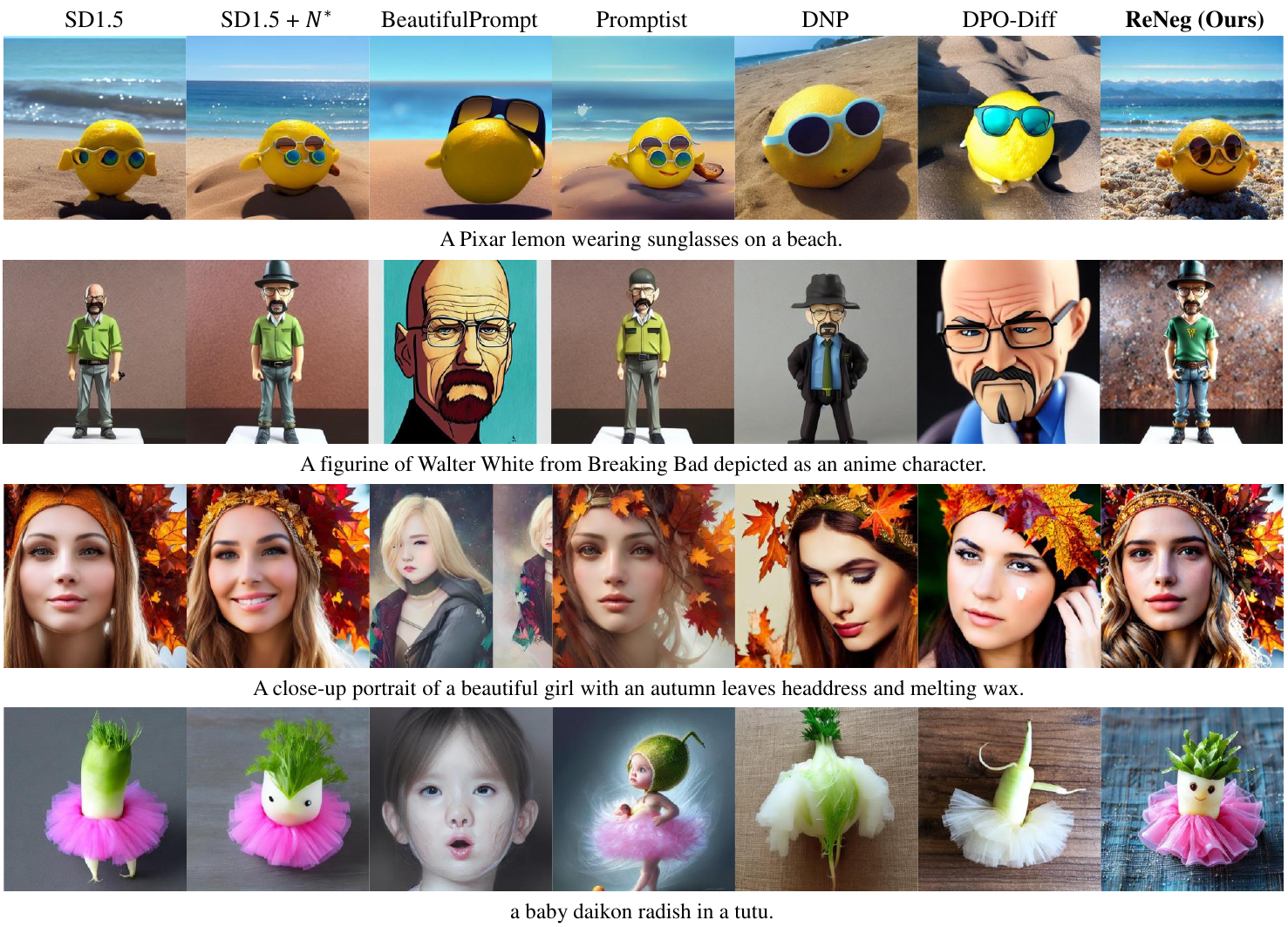}
\end{center}
\vspace{-0.5cm}
   \caption{Qualitative comparisons. The prompts source from HPSv2 and Parti-Prompts benchmarks. All images are generated at a resolution of $512 \times 512$. For a fair comparison, the images above are generated using the same initial noise and seed.}
\label{fig:qualitative_result}
\end{figure*}

\noindent
\textbf{Quantitative results.} First, we compare the proposed method with a handcrafted negative prompts, which consolidates commonly suggested negative prompts from the community. By incorporating the handcrafted negative prompts, SD1.5 achieves performance gains on both benchmarks. However, it still lags behind our optimized negative embedding. We further compare our method with prompt optimization~(recaptioning) methods and outperform them by a large margin. Remarkably, our method achieves comparable performance against the TextCraftor, which involves full finetuning of the UNet weights. The quantitative results highlight that our method not only enhances the visual appeal of generated results but also aligns more closely with human preferences. 
Notably, our method can achieve additional performance gains by combining with positive prompt refinement. Moreover, adaptively finetuning the negative embedding for a given positive prompts leads to further improvements in generation quality.

\noindent
\textbf{Qualitative results.}
The qualitative comparisons are presented in Fig.~\ref{fig:qualitative_result}. By incorporating a handcrafted negative prompts, SD1.5 achieves significant improvements in both visual quality and semantical reasonability. Compared to that, the methods for positive prompt refinement~\cite{beautifulprompt,promptist} yield minimal gains in visual quality. This limited improvement can be attributed primarily to the fact that the optimized prompts generated by these methods sometimes deviate from the intended meaning of the original description, resulting in outputs that do not fully capture the desired content.
Images generated by DNP~\cite{dnp} and DPO-Diff~\cite{dpo-diff} lack finer details and aesthetic appeal. This is partially because of their constrained search strategies and search space, which pose a challenge to finding an optimal negative prompts. Our method, however, produces higher-quality images with finer details and coherent semantic alignment to the text prompts.

\subsection{Visualization of Negative Embedding}
\begin{figure}[t]
\begin{center}
\includegraphics[width=1\linewidth]{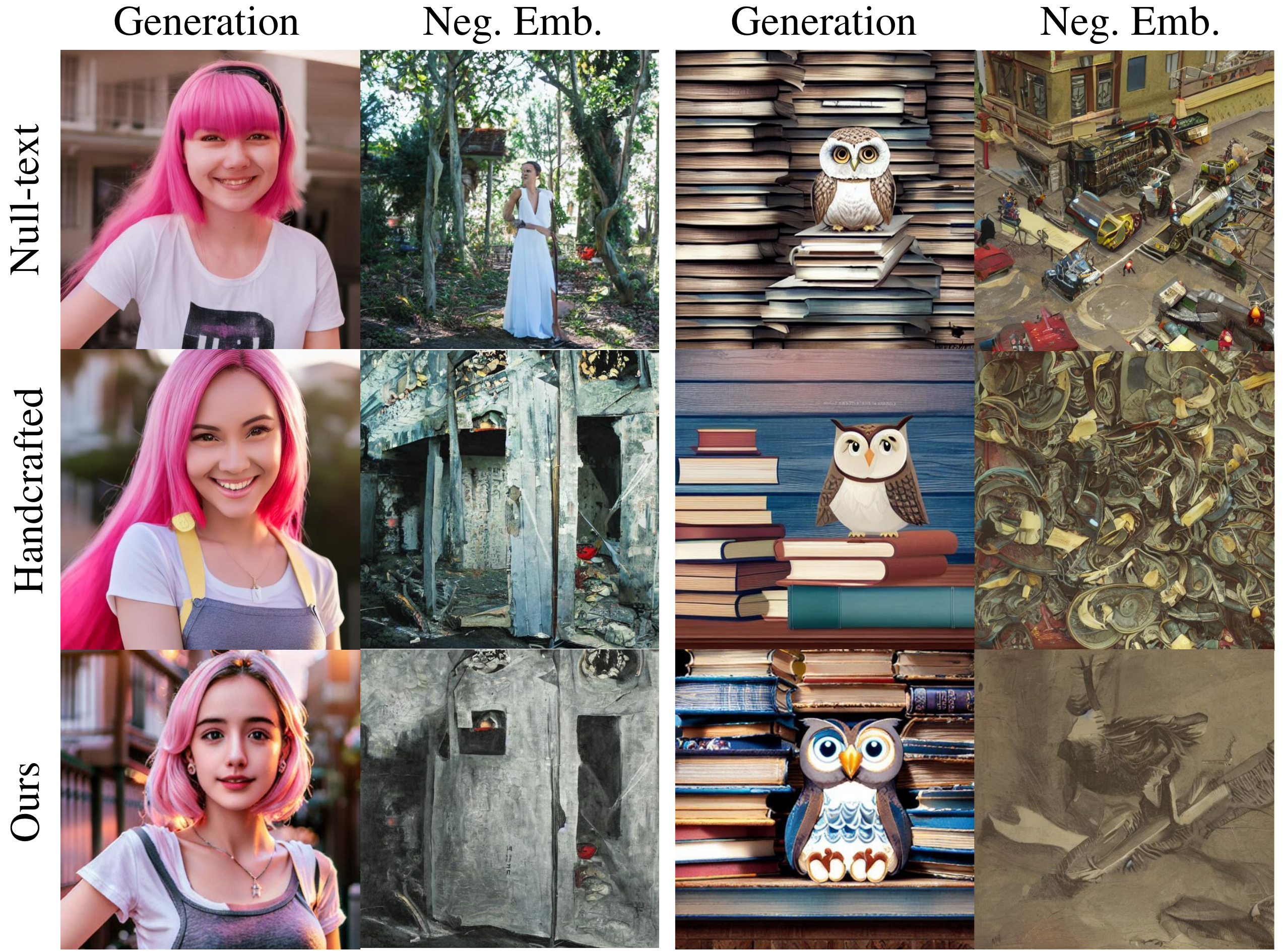}
\end{center}
    \vspace{-0.5cm}
   \caption{Example of generation results using different negative embeddings for the same prompt and the corresponding negative embeddings visualization. \textbf{Prompts: (Left)} \emph{Frontal portrait of anime girl with pink hair wearing white t-shirt and smiling.} \textbf{(Right)} \emph{A plushy tired owl sits on a pile of antique books in a humorous illustration.}}
\label{fig:vis-explain}
\end{figure}

To elucidate the differences among our learned negative embedding, the null-text prompt, and the handcrafted negative prompts, we visualize their corresponding embeddings. Utilizing SD1.5 as the foundational model, we treat negative embeddings as pseudo-positive prompts to generate corresponding images. The observed variations in appearance between the two columns of negative embeddings arise from differing random noise. As illustrated in Fig.~\ref{fig:vis-explain}, our learned embedding exhibits more muted colors and lacks significant semantic information compared to both null-text and handcrafted prompts. Specifically, the null-text prompt represents an average distribution of naturally generated images, closely aligning with authentic visual data. In contrast, the handcrafted prompts displays slight deviations from natural images; while it abstracts certain features, it retains recognizable semantic elements, such as textures and discernible semantic information. Conversely, the images generated from our learned embedding appear atypical and lack clear semantic content. Notably, the outputs from our method align more closely with human aesthetic preferences and maintain semantic coherence.

\begin{table}[t]
\renewcommand\arraystretch{1.1}
\centering
\caption{
Automated win rates on the HPSv2 benchmark comparing SD1.4, SD1.5, and SD2.1 using our negative embedding versus a handcrafted negative prompt. Results are calculated based on the HPSv2.1 metric.
}
\vspace{-0.2cm}
\label{tab:adaption_to_t2i_models}
\resizebox{0.40\textwidth}{!}{%
\fontsize{26pt}{30pt}\selectfont
\begin{tabular}{lccccc}
\toprule
\textbf{Model} & \textbf{Animation}  &  \textbf{Concept Art}  & \textbf{Painting}  & \textbf{Photo} & \textbf{Average}  \\ 
\midrule
SD1.4 & 0.91 & 0.95 & 0.97 & 0.87 & 0.93 \\
SD1.5 & 0.99 & 0.99 & 0.98 & 0.99 & 0.99 \\
SD2.1 & 0.96 & 0.98 & 0.98 & 0.93 & 0.96  \\
\bottomrule
\end{tabular}
}
\end{table}

\begin{table}[t]
\renewcommand\arraystretch{1.1}
\centering
\caption{
Performance comparison on video generation models. ‘Handcrafted Prompt’ denotes the handcrafted negative prompts.
}
\label{tab:adaption_to_t2v_models}
\resizebox{0.48\textwidth}{!}{%
\fontsize{26pt}{30pt}\selectfont
\begin{tabular}{lccccc}
\toprule
\textbf{Model} & \textbf{\makecell{Aesthetic \\ Quality}} & \textbf{\makecell{Motion \\ Smoothness}} & \textbf{\makecell{Temporal \\ Flickering}}  & \textbf{\makecell{Background \\ Consistency}} \\ 
\midrule
VideoCrafter2 & 58.0 & 97.7 &96.2 & 97.6\\
w/ Handcrafted Prompt & 57.8 & 97.8 & \textbf{96.5} &97.9 \\
w/ Our ReNeg & \textbf{58.6} & \textbf{97.8} & 96.4 & \textbf{98.5}\\
\midrule
ZeroScope & 49.9 & \textbf{98.9} & \textbf{98.3} & 97.7 \\
w/ Handcrafted Prompt & 53.1 & 98.6  & 97.9 & 98.2 \\
w/ Our ReNeg & \textbf{58.7} & 98.1 & 97.3 & \textbf{98.7}  \\
\bottomrule
\end{tabular}
}
\end{table}

\subsection{Generalization Capability}
\begin{figure}[t]
\begin{center}
\includegraphics[width=1\linewidth]{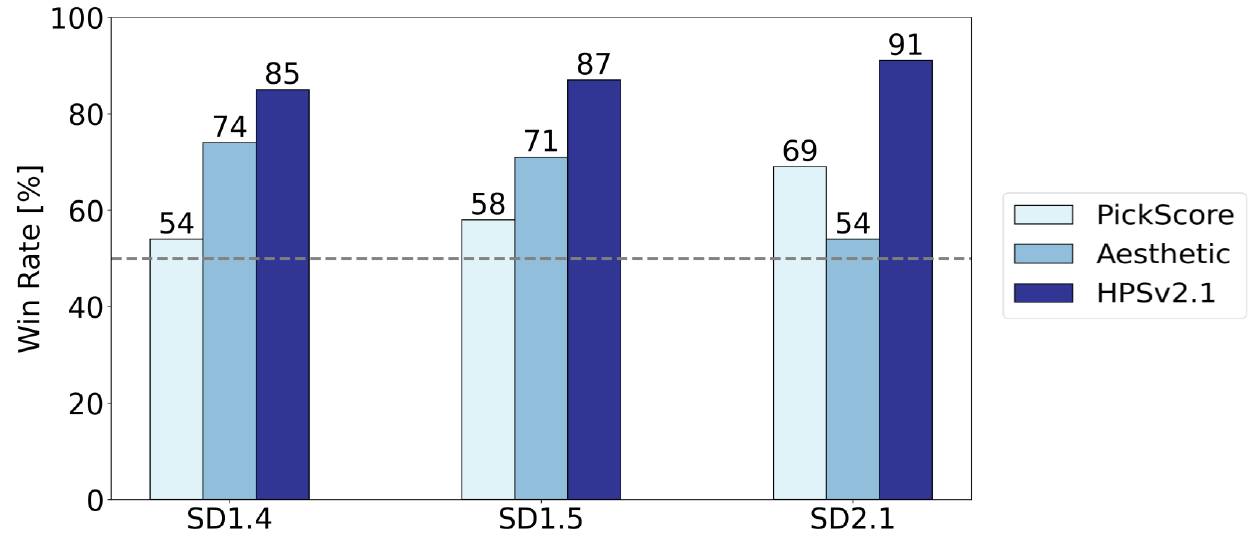}
\end{center}
   \vspace{-0.4cm}
   \caption{
   Comparison of the win rates on Parti-Prompts with and without our ReNeg. Across three metrics, our negative embedding achieves strong performance on various T2I models.
   }
\label{fig:win_rate_parti}
\end{figure}

\noindent
\textbf{Generalization across different SD models.}
The proposed ReNeg is flexible and can be easily applied to various SD models, including SD1.4 and SD2.1. We calculate the win rate between the SD models using handcrafted negative prompts and the learned negative embedding on HPSv2 and Parti-Prompts.  
To provide a more meaningful comparison, we calculate the win rate of using our negative embedding relative to the handcrafted negative prompts, rather than comparing it to a setup that exclusively uses a positive prompts (\textit{i.e.,} without any negative prompts).
As shown in Tab.~\ref{tab:adaption_to_t2i_models} and Fig.~\ref{fig:win_rate_parti}, the learned negative embedding significantly outperforms the handcrafted counterpart, achieving substantial improvements in human preference alignment.

\begin{table}[h]
\renewcommand\arraystretch{1.1}
\centering
\caption{
Performance comparison on ControlNet. 
Our negative embedding enables a more visually appealing pose-to-image effect and better aligns with human preferences.
}
\label{tab:quantitative_controlnet}
\resizebox{0.40\textwidth}{!}{%
\fontsize{26pt}{30pt}\selectfont
\begin{tabular}{lccccc}
\toprule
\textbf{Model} & \textbf{PickScore}  &  \textbf{Aesthetic}  & \textbf{HPSv2.1}  \\ 
\midrule
ControlNet & 19.49 & 5.54 & 26.83  \\
w/ Handcrafted Prompt & \textbf{19.54} & 5.54 & 27.60  \\
w/ Our ReNeg & 19.52 & \textbf{5.95} & \textbf{30.79}   \\
\bottomrule
\end{tabular}
}
\end{table}

\noindent
\textbf{Generalization across T2I and T2V Models.}
Here, we conduct experiments to examine the generalization capability of the proposed method across different generative models and tasks. As shown in Fig.~\ref{fig:teaser}, the learned negative embedding can be transferred seamlessly to the text-to-image and text-to-video models, including ControlNet, ZeroScope, and VideoCrafter2. The quantitative results are reported in Tab.~\ref{tab:adaption_to_t2v_models} and Tab.~\ref{tab:quantitative_controlnet}. We observe a consistent performance improvement on both ControlNet and T2V models by incorporating the learned negative embeddings, which reveal strong generalization capabilities of the proposed ReNeg. To conclude, the learned negative embedding can be easily shared across any text-conditioned generative models using the same text encoder. More qualitative results can be found in the appendix.

%% file: sec/5_conclusion.tex
\section{Conclusion}
\label{sec:conclusion}
We propose~\method,  a framework that searches for a global negative embedding under reward feedback. Building on this, we adaptively learn a distinct negative embedding tailored to each positive prompt, which exhibits consistent improvements in detail refinement and textual alignment. To enable effective gradient propagation through reward guidance, we incorporate the CFG-training strategy. Despite its simplicity, our negative embedding proves to be highly useful, surpassing results generated with only positive prompts or handcrafted negative prompts and rivaling those achieved through full finetuning of diffusion models. Moreover, our negative embedding can be easily transferred to other T2I or T2V models, provided they share the same text encoder.
Although~\method can generate visually appealing images, limited by the generative ability of the base model, it sometimes shows semantic deviations from the prompt. The improvement is left to our subsequent work.